\documentclass[10pt,conference]{IEEEtran}
\usepackage{cite}
\usepackage{amsmath,amssymb,amsfonts}
\usepackage{algorithm} 
\usepackage{algpseudocode} 
\usepackage{graphicx}
\usepackage{textcomp}
\usepackage{xcolor}
\usepackage{float}

\usepackage{wrapfig}
\usepackage{tabularx}
\usepackage{multirow}
\usepackage{hyperref}
\usepackage[english]{babel}
\usepackage{todonotes}
\usepackage{enumerate}
\usepackage{enumitem}
\usepackage{comment}
\usepackage{ragged2e}
\usepackage{soul}
\usepackage{booktabs}

\usepackage{listings}

\def\BibTeX{{\rm B\kern-.05em{\sc i\kern-.025em b}\kern-.08em
    T\kern-.1667em\lower.7ex\hbox{E}\kern-.125emX}}

\begin{document}

\title{A Reliable and Low Latency Synchronizing Middleware for Co-simulation of a Heterogeneous Multi-Robot Systems}

\author{\IEEEauthorblockN{Emon Dey$^{\dagger}$, Mikolaj Walczak$^\S$, Mohammad Saeid Anwar$^{\dagger}$, Nirmalya Roy$^{\dagger}$}
\IEEEauthorblockA{\textit{$^\dagger$Mobile Pervasive \& Sensor Computing Lab, Center for Real-time Distributed Sensing and Autonomy (CARDS)}\\ \textit{$^\dagger$Department of Information Systems, University of Maryland, Baltimore County, USA} \\
\textit{$^\S$Department of Computer Science and Electrical Engineering, University of Maryland, Baltimore County, USA} \\
\{edey1, mwalcza1, wy71697, nroy\}@umbc.edu
}
}

\maketitle
\thispagestyle{plain}
\pagestyle{plain}

\begin{abstract}
Recent Internet-of-Things (IoT) networks span across a multitude of stationary and robotic devices namely unmanned ground vehicles, surface vessels, and aerial drones to carry out mission-critical services such as search and rescue operations, wildfire monitoring, flood/hurricane impact assessment. Achieving communication synchrony, reliability, and minimal communication jitter among these devices is a key challenge both at the simulation and system levels implementation due to the underpinning differences in physics-based robot operating system (ROS) simulator that is time-based and network-based wireless simulator that is event-based, in addition to the complex dynamics of mobile and heterogeneous IoT devices deployed in real environment. Nevertheless, synchronization between physics (robotics) and network simulators is one of the most difficult issues to address in simulating a heterogeneous multi-robot system before transitioning it into practice. The existing TCP/IP communication protocol-based synchronizing middlewares mostly relied on robot operating system 1 (ROS1) which expends a significant portion of communication bandwidth and time due to its master-based architecture. To address these issues, we design a novel synchronizing middleware between robotics and traditional wireless network simulators relying on the newly released real-time ROS2 architecture with a masterless packet discovery mechanism. We propose a ground and aerial agents' velocity-aware Transmission Control Protocol (TCP) algorithm using the publish-subscribe transport of Data Distribution Service (DDS) to minimize the packet loss and synchronization, transmission, and communication jitters between a diverse set of robotic agents. Our proposed middleware is agnostic to specific robotics and network simulators, but for simulations and experiments, we employ Gazebo as a Physics-based ROS simulator and NS-3 as a wireless network simulator. We performed extensive network performance evaluations both at the simulation and system levels in terms of packet loss probability and average latency with line-of-sight (LOS)/non-line-of-sight (NLOS) and TCP/UDP communication protocols over our proposed ROS2-based synchronization middleware. Moreover, for a comparative study, we presented a detailed ablation study replacing NS-3 with a real-time wireless network simulator, EMANE, and masterless ROS2 with master-based ROS1. Finally, to make the transition in practice, we deployed a diverse set of real robots -- one aerial drone (Duckiedrone) and two ground vehicles (TurtleBot3 Burger) in different terrains, forming both masterless (ROS2) and master-enabled (ROS1) clusters to evaluate potential network synchronization and jitter issues. Our proposed middleware attests to the promise of building a large-scale IoT infrastructure with a diverse set of stationary and robotic devices achieving low-latency communications (12\% and 11\% reduction in simulation and real) while satisfying the reliability (10\% and 15\% packet loss reduction in simulation and real) and high-fidelity requirements of mission-critical applications.
\end{abstract}

\section{Introduction}
In the field of robotics and wireless sensor networks, multi-agent systems play a vital role due to their ability in forming large, interconnected networks with coordination among agents that make them an integral part of a variety of robotic and smart city IoT applications \cite{gomes2018co,Taemin2018IOT}. For example, Unmanned Aerial Vehicle (UAV) systems are being increasingly used in a broad range of applications requiring extensive communications, either to collaborate with other UAVs \cite{kudelski2013robonetsim, brunori2021reinforcement} or with Unmanned Ground Vehicles (UGV) \cite{Phillip2012Anvel} and WSNs \cite{seah2009wireless, chen2021data}. Particularly in a smart city environment, the presence of heterogeneous wireless sensor networks, aerial and ground vehicles coordinating with each other is going to be an essential feature in the future. The mutual information transfer between aerial drones, ground robots and IoT networks can help to make intelligent decisions in a disaster-prone area in critical time.\\
Synchronized communications between UAVs, UGVs and IoT networks can aid in spurring situation awareness in the smart environments among each deployed robotic asset and IoT device, seamlessly executing commands for civilian applications, and meeting the QoS requirements for high-fidelity military applications. Deploying such systems directly in the actual environment may bring in harmful consequences as they necessitate the extensive fine-tuning of algorithm parameters \cite{Rosnetsim2021}. Therefore, it is essential to simulate such a cross-cutting robotic and IoT system a-priori using appropriate simulation toolkits in order to pinpoint and solve the underlying research and implementation challenges. The existing research \cite{Rosnetsim2021} has focused mainly on simulating such an environment among multiple agents i.e., two aerial systems to pose the communication challenges between homogeneous robots using robot operating systems 1 (ROS1). While ROS1 was the defacto standard for robotics middleware, to meet the requirements of real-time distributed embedded systems, ROS2 has been recently introduced with its Data Distribution Service (DDS) capability~\cite{Schlesselman2004OMGDS,PardoCastellote2003OMGDS}. DDS is an industry-standard real-time communication system and has been shown to meet the requirements of distributed systems for resilience, fault-tolerance, security and scalability~\cite{Xiong2006EvaluatingTP,Sierla2003}. DDS with underlying masterless publish subscribe data transport architecture in ROS2 guarantees the requirements of real-time distributed embedded systems by being scalable to various operating systems and amenable to various transport layer configurations, such as deadline, fault-tolerance and process synchronization~\cite{maruyama2016exploring}. 

Nonetheless, other than the aforementioned constraints in ROS1 due its master-based publish subscribe architecture. In contrast to masterless packet discovery mechanism in ROS2 for satisfying the real-time distributed embedded systems requirements, the main bottleneck in this endeavor lies in the necessity of synchronizing two different simulators which have disparate operating principles \cite{moon2020gazebo,fields2016simulation}. One of these is physics simulators that account for replicating the interaction between physical robots and their operating environments. The other is network simulators that estimate the deployed agents’ communication performance over the wireless network. Therefore, one of our key objectives is to identify and evaluate the underpinning communication challenges between a Physics-based ROS simulator toolkit that is time-based, and a Network-based simulator that is event based. For building a resilient and effective Perception-Action-Communication framework focusing on simulating jointly the interactions of physical environments, with network traffic by integrating the real-time features of ROS2 with masterless packet discovery mechanism. The major challenges to realize such a seamless networking infrastructure cross-cutting between robotics platforms and IoT, are: effectively representing the interactions of robotic assets with the physical environments and synchronizing the packet-level communications between multiple agents. This will help to accurately model the simulated behavior representing the reality and satisfy the various high fidelity robotic and IoT applications in challenged environments.

Needless to say, building multi-agent systems is strenuous. Therefore, resilient and coordinated communications across heterogeneous simulated and real systems can make our proposed middleware framework less error-prone and more accurate, reliable, fault tolerant and inter-operable. For example, SynchroSim \cite{dey2022synchrosim}, FlyNetSim \cite{baidya2018flynetsim}, ROS-NetSim \cite{calvo2021ros}, CORNET \cite{acharya2020cornet}, CPS-Sim \cite{suzuki2018cps}, and RoboNetSim \cite{kudelski2013robonetsim} are some of such kinds of works in existing literature that have been implemented in ROS1 using AirSim \cite{shah2018airsim}, ARGoS \cite{Carlo2012Argos}, Gazebo \cite{Gazebo} as physics simulators and OMNeT++ \cite{varga2010omnet++}, NS-2 \cite{NS-2}, NS-3 \cite{Riley2010NS-3}, Mininet \cite{Mininet} as network simulators. Nonetheless, some of the major drawbacks that exist in those works are: {\em (i)} the proposed systems are compatible with either UAVs or UGVs, but not with heterogeneous agents such as a mix of both UAVs and UGVs under the same network, {\em (ii)} it suffers from low co-simulation speed, {\em (iii)} it gives rise to floating-point arithmetic error during the time synchronization, {\em (iv)} it is difficult to set up proper window size for reliable packet transmissions in presence of diverse agents with different velocities. {\em (v)} it loses synchronization when the relative speed varies between the simulators, and {\em (vi)} the network does not self-heal or reconfigure to maintain seamless and uninterrupted connections in presence of node loss or adversarial attacks.

Notably there are real-time network emulators, as an exemplar, EMANE \cite{ahrenholz2011integration} which can help solve some of the aforementioned issues, but we empirically observed that it has a comparatively higher packet delay and is not quite matured enough to be integrated with Robot Operating Systems (ROS). Motivated by this, we propose a novel synchronizing middleware that follows the publish-subscribe transport of Data Distribution Service (DDS) architecture~\cite{maruyama2016exploring} and the TCP/IP protocol in order to facilitate the development of an appropriate ROS2-compatible synchronizing middleware to overcome some of the specific challenges as listed above. To the best of our knowledge, it is the first endeavor to develop a synchronizing middleware for a real-time masterless ROS2 environment. The major advantage of publish-subscribe architecture in ROS2 over ROS1 is that during the process of transmitting and exchanging data between multiple agents, ROS2 takes into account the total number of agents that have been deployed within a particular cluster and chooses the sliding window that is optimal for that data transmission. We employed Gazebo, a Physics-based robotic simulator and NS-3, a discrete event network simulator for modeling and co-simulating jointly the interactions of physical environments with the network traffic. NS-3 depicts the process as a distinct sequence of time occurrences, as opposed to Gazebo, which uses the system clock of the machine as the simulation time. We postulate a velocity-aware dynamic window-based Transmission Control Protocol (TCP) algorithm in presence of heterogeneous ground and aerial agents using the publish-subscribe transport of Data Distribution Service (DDS) of ROS2 environment to minimize the average packet loss and latency. We evaluate the performance of our proposed algorithm after deploying it on a cluster that is comprised of real-world UAVs – Duckiedrone and UGVs --  TurtleBot. The specific research contributions are summarized below.

\begin{itemize}

\item {\bf Reliable and faster synchronization middleware for co-simulation of multi-agent systems:} We design a robotic and wireless network simulators agnostic co-simulation middleware for heterogeneous multi-agent communications. We depict a customized version of the Transmission Control Protocol (TCP) using the publish-subscribe transport of Data Distribution Service (DDS). We implement our TCP algorithm using real-time robotic operating system 2 (ROS2), namely, {\em Foxy} which supports masterless packet discovery and real-time requirements of distributed embedded systems. We leverage the low latency virtue of ROS2 along with the TCP/IP protocol to improve the reliability of multi-agent communication paradigm with our proposed customized sliding window-based TCP algorithm. Our proposed transmission control algorithm can vary the window size dynamically while considering the velocity differences between the ground and aerial agents in a heterogeneous setting to minimize the process synchronization and communication delays.

\item {\bf System implementation using off-the-shelf real robotic devices:} We extend our simulation based synchronization middleware framework into the real world robots to witness the various network performance metrics such as probability of packet loss and average network latency in case of actual deployment. We assemble commercially available UASs and UGVs robots -- Duckiedrone and TurtleBot3 Burger and form a cluster using one UAS and two UGVs to implement our proposed multi-agent synchronizing middleware and evaluate the communication performance. In particular, we record \textit{ROSbag} of uniform size containing information regarding the physical robot states in the wild and send it between the other robotic agents to record the network performance metrics.

\item {\bf Empirical evaluation considering different communication scenarios:} We perform extensive performance evaluations, taking into account both line-of-sight (LOS) and non-line-of-sight (NLOS) communication scenarios on the basis of probability of packet loss, and average packet delay. We employ Gazebo as Physics simulator and NS-3 as network simulator at the simulation level to report the network synchronization and performance results. We also perform an ablation study with a real-time network simulator \textit{EMANE} and compare our proposed masterless ROS2 synchronization middleware  with master-based ROS1 synchronization middleware. Experimental results both at the simulation and system levels attest that our proposed synchronizing middleware in ROS2 integrated with Transmission Control protocol (TCP) outperforms the traditional ROS1 systems in terms of ensuring fewer packet losses (10\% reduction in simulation and 15\% in real) and faster data transmissions rates (12\% increase in simulation and 11\% in real).

\end{itemize}

\section{Challenges} \label{challenge}
\begin{figure}[htbp]
\centering
\includegraphics[width=\linewidth]{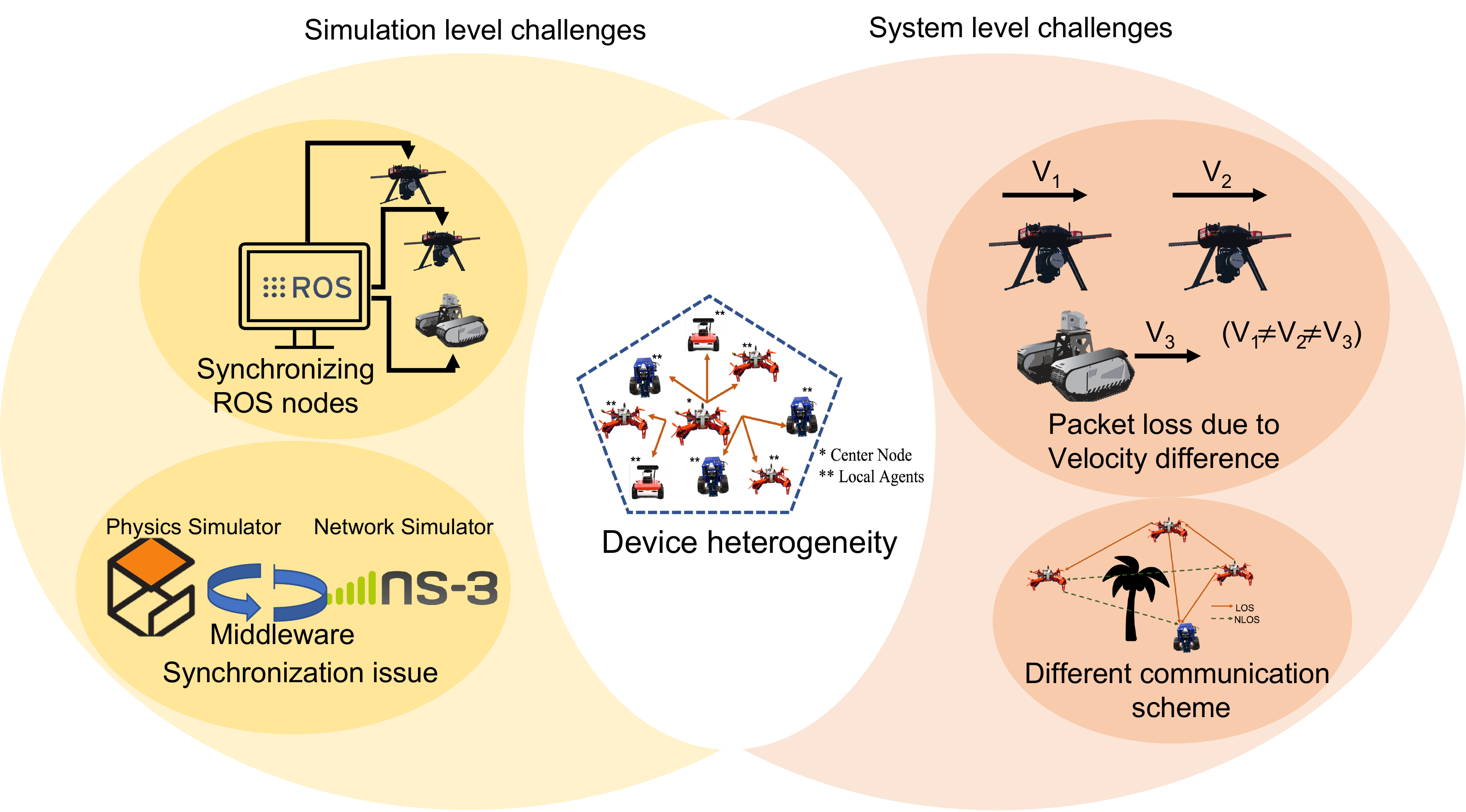}
\caption{Illustration of the possible hardware and software centric issues while working with multi-agent robotic implementation.}
\label{issue}
\vspace{-2ex}
\end{figure}

Implementation of a multi-agent heterogeneous system raises a number of challenges due to its inherent nature. The need for prior simulation to get an estimation of the actual performance adds specific software-centric challenges along with deployment issues. The challenges for these two cases are identified separately in figure \ref{issue}. The left lobe of the figure signifies the simulation level challenges. The two sub-modules inside of it are the two most fundamental research questions in simulating multi-agent systems and working with simulators with different paradigms. When measuring the communication performance among the agents in a simulated environment, the integration of a physics simulator and a network simulator is necessary. However, the issue is that the physics simulator gets updated over time, and the network simulator works based on available events. So, to run those two simulators synchronously (co-simulation), a synchronizing middleware is needed. Also, if heterogeneity exists among the robotic agents, it becomes even more complex to simulate such a scenario. The specification disparity and different capabilities of data packet handling are some of the major reasons for that. 
Additionally, during the deployment with actual robotic agents, if the velocity difference between agents exceeds a certain threshold, it can cause severe degradation in the communication performance. Also, the presence of obstacles between the communicating agents (non-line-of-sight scenario) can be considered as another hindrance for the deployment point-of-view.

\section{Background and Related Work} \label{sec:related_work}
In this section, we will provide a quick overview of the concepts we have accepted as well as the work that has been completed in regard to the various aspects of our system. 
\subsection{Simulation Tools}
Previously published studies that couple multiple simulators PiccSim \cite{Piccsim}, NCSWT \cite{NCSWT}, ModelSim \cite{Modelsim}, used MATLAB/Simulink and NS-2 as physics and network simulators. For example, the ARGoS physics simulator and the NS-2 and NS-3 network simulators were used in some UAV based studies such as RoboNetSim \cite{kudelski2013robonetsim}. The control dynamics and networking system of an NCS can be accurately simulated using other co-simulation platforms such as adevs + NS-2 \cite{Adevs}, Modelica + NS-2 \cite{Modelica}, and Matlab/Simulink + OPNET \cite{Opnet}. Integration between different co-simulation frameworks (DEVS) is based on either standardizing on the High Level Architecture (HLA) or the Discrete Event System Specification (DESS).\\ 

Additionally, in terms of physics simulators, AirSim\cite{shah2018airsim}, and ANVEL\cite{durst2012real, fields2016simulation} provides such a toolkit by combining popular graphical representation methods. While both AirSim and ANVEL have significant simulation capabilities, they struggle to create large-scale complex visually rich environments that are more realistic in their representation of the real world, and they have fallen behind various advancements in rendering techniques made by platforms such as Unreal Engine or Unity\cite{Usimulator}. Another such stable simulator is Gazebo\cite{moon2020gazebo} has a modular design that allows for the implementation of various physics engines, sensor models, and the development of 3D worlds.  
In terms of network simulators, OMNeT++\cite{varga2010omnet++} is an object-oriented and modular discrete event network simulation framework. OMNeT++ also supports parallel distributed simulation, and inter-participant communications can be achieved via a variety of approaches. However, we chose to utilize event-based simulator, NS-3\cite{Riley2010NS-3}, in which the scheduler normally runs the events sequentially without synchronization with an external clock. The NS-3 simulator shows all of the network models that make up a computer network and also the features offered by this simulator greatly overlaps with our requirement.

\begin{table*}[h!]
\centering
\caption{Summary of Prior works on developing synchronizing middleware between physics and network simulators.}
\label{prior}
\resizebox{\linewidth}{!}{%
\begin{tabular}{|l|l|l|}
\hline
\textbf{Synchronization Method} & \textbf{Working Procedure} & \textbf{Issues} \\ \hline
Time-stepped \cite{Piccsim, kudelski2013robonetsim} & \begin{tabular}[c]{@{}l@{}}Used advanced network simulation tools together with robotic simulators.\\ At predefined thresholds, the simulators halt their individual simulations \\ to share data with each other.\end{tabular} & \begin{tabular}[c]{@{}l@{}}The network simulator isn’t capable to take into \\ account the presence of obstacles.\\ Data exchange issues.\end{tabular} \\ \hline
Time-stepped with scheduler \cite{baidya2018flynetsim} & Common sampling period to be used by both simulators. & \begin{tabular}[c]{@{}l@{}}If the network simulator runs faster than the \\ physics simulator, the network events must be \\ buffered in a cache and wait to be processed until \\ the next sampling time.\end{tabular} \\ \hline
Variable-stepped \cite{acharya2020cornet} & \begin{tabular}[c]{@{}l@{}}A global event scheduler keeps track of all the events from both \\ simulators and schedules based on their timestamps, \\ allowing only one event process to operate at a time.\end{tabular} & Float-point arithmetic error. \\ \hline
Global event driven \cite{suzuki2018cps} & The server is forced to reproduce the exact same time-steps. & Higher latency during co-simulation. \\ \hline
Sliding window \cite{dey2022synchrosim, Rosnetsim2021} & \begin{tabular}[c]{@{}l@{}}Capture and track network events over the window period and allow the \\ network simulator to step up to the end of the window.\end{tabular} & \begin{tabular}[c]{@{}l@{}}Difficulty in maintaining performance for \\ multi-agent system with disparate velocity.\end{tabular} \\ \hline
\end{tabular}%
}
\end{table*}

\subsection{Synchronizing Middleware}
The working principles of existing middleware for the synchronization purpose varies from time-stepped to sliding window based protocol. In some works, a scheduler was also introduced to decrease latency and a improved version of it named variable-stepped was adopted. In table \ref{prior}, we have summarized such existing approaches along with the probable issues may arise if we choose to utilize them in our use case. 
Another aspect in this topic is the choice of communication protocol for middleware testing. RUDP \cite{RUDP} provides a solution for real-time embedded systems with transmission speed and reliability criteria that TCP and UDP have not been able to achieve. In a real-world demonstration, the claim that the suggested technique can give a faster throughput than TCP without experiencing packet loss has been validated. But the issue with RUDP is that, Robot Operating System hasn't yet finalized the compatibility test with it. In this work, UAV and UGV components are being simulated with 3D visualization using Gazebo, while the network infrastructure is being provided by NS-3 and middleware is being developed for the creation of an inter-simulation data-path with time and position synchronization at both ends using our co-Simulation of robotic networks. The whole operation will be conducted under a ROS2 environment.

\section{Methodology} \label{method}
The design basics of our proposed synchronizing middleware, modification of conventional TCP/IP architecture, integration procedure with masterless DDS approach are described in the section.

\subsection{Synchronization Middleware Design through Modifying the TCP/IP Architecture}
The sliding window-based synchronization strategy \cite{dey2022synchrosim} is the foundation of our algorithm design for the master-less multi-agent scenario. When dealing with a multi-agent system, the choice of window size is the most important design parameter that you can choose. In such a case, a fixed window size at all points will not be relevant since different agents may perform better with different radio frequencies, and there may also be differences in terms of the hardware specifications. Moreover, there may be differences in the amount of available memory. Our system begins with a value for the window size $w$ that has been manually initiated and then takes into account the total number of agents $T$ that are present for a particular scenario. It chooses the appropriate publisher ($P$) and subscriber ($S$) for the data transfer event based on the information provided by both parties. After that, it performs a stringent check on the difference in velocity between the participating agents in order to accurately calculate an appropriate window size for a particular communication event. The process of velocity difference calculation can be found in \ref{algo:1}. When it comes to the transmission scenario between a ground vehicle and an aerial vehicle, this technique becomes more apparent. We decided to use the packet loss probability ($L_p$), and average delay ($PD_a$), which are two of the most important parameters to monitor while sending valuable information, in the operating moment of the middleware in order to keep the level of information that is transmitted at an acceptable point. This was done in order to keep the level of information that is transmitted at a satisfactory point. In order to inherently determine the likelihood of packet loss as well as the average delay and report it within a synchronization window, the following equations are used:\\
Packet loss probability,
\begin{equation}
{ }_{n}^{i} L p=1-\frac{{ }_{n}^{i} D_{s}}{{ }_{n}^{i} D_{p}}
\label{loss}
\end{equation}
Here, \({ }_{k}^{i} D_{s}\) is the number of data packets delivered to the total number of subscribers and \({ }_{n}^{i} D_{p}\) is the number of data packets transmitted from the publishers. The average value is reported after the data transmission is complete between each publisher-subscriber pair.\\

Average packet delay,
\begin{equation}
\textit{$PD_a$} = \textit{$D_{pr}$} + \textit{$D_t$} + \textit{$D_{pg}$} + \textit{$D_q$}
\label{delay}
\end{equation}
Here, \textit{$D_{pr}$} is processing delay \textit{$D_t$} is transmission Delay, and \textit{$D_{pg}$}, \textit{$D_q$} represent propagation, and queuing delay respectively.

\begin{algorithm}
	\caption{Algorithm of synchronizing middleware for masterless environment}
	\begin{algorithmic}[1]
	\State\textbf{Input:} Data packets $D$, Total number of agents $T$, window size $w$, velocity of agents $V$
	\State\textbf{Output:} Packet loss probability $Lp$, and average delay $PD_a$\\
	\textbf{Simulation Initialization:} Publisher $P$ and subscriber $S$ agents ip are determined where $P,S\in T$, initialize physics simulator time = 0, window size $w$ \\
	\textbf{Data transmission and Update co-simulation:} Select topic of interest and transmit data packets from publisher\\
	    \textit{Update} the timestamp t = t
	    \If {any event exist}
	    \State\textbf{Sliding Window adaptation:} Acquire the velocity of Publisher $V_p$ and Subscriber $V_s$ in meters/sec\\
    	Calculate the required adaptation of sliding window, \begin{equation}
w_{a}=w+\left(V_{p}-V_{s}\right) / 1000
\end{equation}
	    \State\textbf{Report $PD_a$ and $L_p$ upon synchronization:} Calculate average delay and packet loss probability for the synchronized event using equation \ref{delay} and \ref{loss} respectively
	    \State\textbf{Timestamp update:} \textit{Update} $t = t + w_a$ and request for next window to iterate the process
	    \EndIf
	\end{algorithmic} 
    \label{algo:1}
\end{algorithm}

\textbf{Updating the co-simulation steps.} Another essential part of our process is the development of a mutual information update method for both the physics simulator and the network simulator. At the outset of the simulation, the Physics simulator establishes two essential characteristics of the agents that will be employed for a particular communication round. These characteristics are the velocity of the agents and the distance that separates them as determined by positional coordinates. The information that the network simulator stores includes the communication scheme that will be used (for reasons related to reliability, we have decided to use a TCP/IP-based approach rather than an UDP-based approach), the number of packets, the length of each packet, and the IP addresses of both the publisher and subscriber agents. After that, the Physics simulator and the network simulator are iterated upon while using a single initially fixed window size. When the simulation is advanced, the Physics simulator sends the information about distance and velocity to the network simulator, and the network simulator uses that information to compute the chance of packet loss. When the value of the loss is compared to a predetermined threshold, the result is sent to the display. In the event that this is not the case, the information is sent back to the synchronizing module, where it is processed, and the initial window size is modified in accordance with the method \ref{algo:1}. This process will continue until a result is obtained for this particular round that is deemed to be satisfactory, after which the initialization procedure will begin once again with a fresh group of agents.

\subsection{Deploying Synchronization Middleware on DDS using ROSbridge}
\begin{figure}[htbp]
\centering
\includegraphics[width=\linewidth]{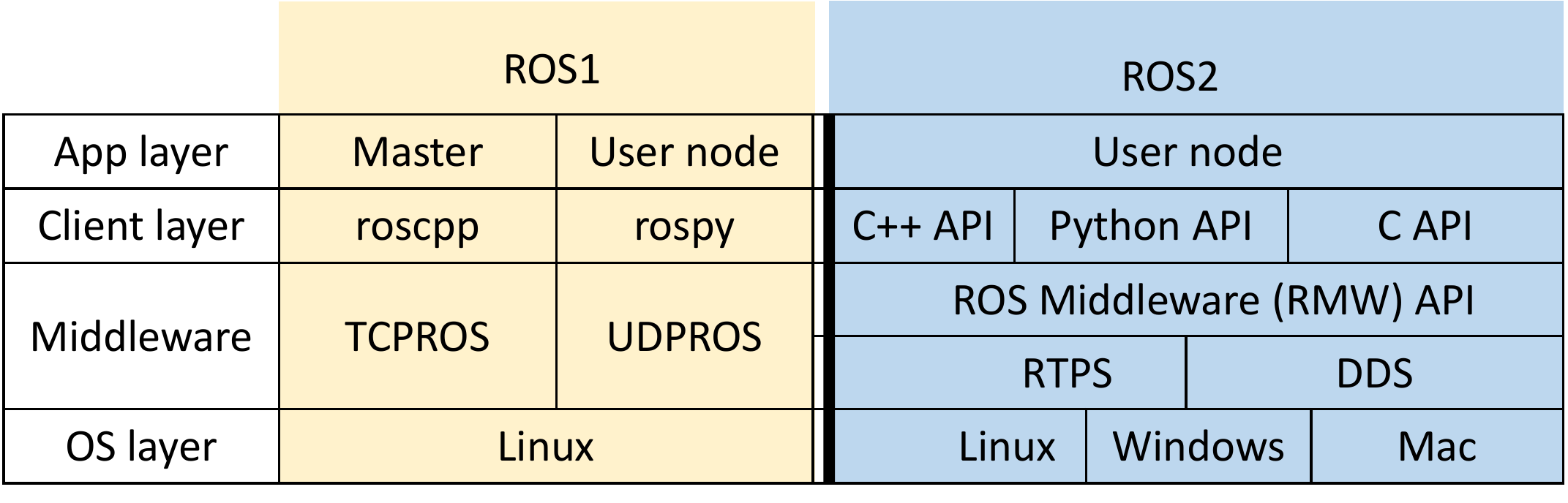}
\caption{Comparison between the two ROS versions.}
\label{ros}
\vspace{-2ex}
\end{figure}

\begin{figure}[htbp]
\centering
\includegraphics[width=\linewidth]{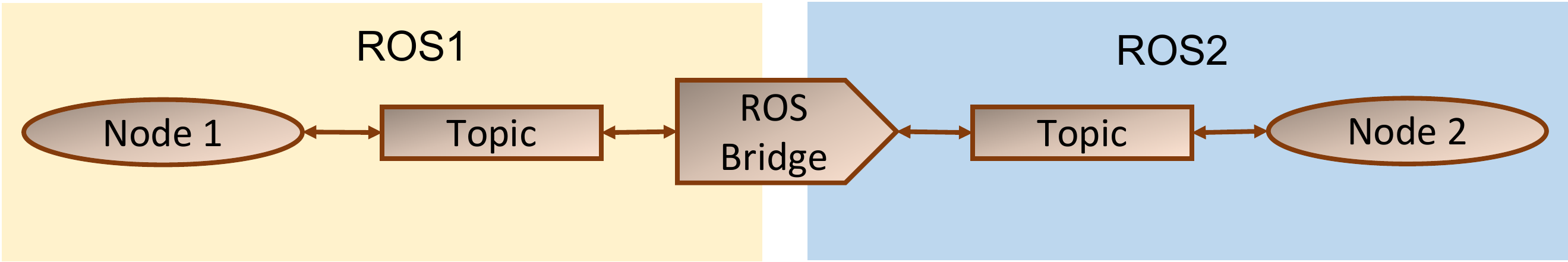}
\caption{ROSbridge implementation to import the ROS1 topics to ROS2 environment.}
\label{bridge}
\vspace{-2ex}
\end{figure}

We use the ROSbridge idea to combine the functionalities of Data Distribution Service (DDS) with our suggested middleware solution. To ensure speedier data transmission, the communication sockets are developed in C, which, unlike ROS1, is a well-supported platform in the ROS2 environment. Some of the operational robots in our configuration do not have ROS2 functionality by default. To prevent the time-consuming re-installation, we built a bridge between two ROS versions to exchange rostopics. The subject flow of ROSbridge is depicted in figure \ref{bridge}. We used DDS to further process the rostopics from a communication standpoint. DDS uses a publish-subscribe framework to send and receive data, events, and commands between nodes. Nodes that generate information (publishers) generate ``topics" and ``samples". DDS delivers samples to subscribers who express an interest. DDS is in charge of message addressing, delivery, and flow control. Any node has the ability to publish, subscribe, or both. We were able to simplify distributed application network programming thanks to the DDS publish-subscribe idea. DDS requires less design time for interactions between nodes. It can also add another essential security feature because applications do not require knowledge about the existence or location of other participants. DDS allows users to configure discovery and behavior approaches based on QoS factors. By permitting anonymous message interchange, DDS promotes modular, well-structured programs. Figure \ref{ros} depicts the functionality of DDS within the ROS2 architecture and its differences from the ROS1 design.

 \section{Simulation Setup} \label{simulation}
 As stated in earlier sections, we have chosen Gazebo and NS-3 as physics and network simulators respectively for our experimentation. Here, we provide the brief description of the simulators, detailed simulation setup and analysis on the achieved simulation results.
 
\textbf{Gazebo.} The fully open-source physics simulator Gazebo is often nicknamed a robotics simulator due to its vast libraries of robot models, sensors, and compatibility with ROS. Being integrated with the ODE physics engine and its ability to be integrated with several other physics engines, allows users to create virtual environments with accurate real-world properties such as: gravity, friction and drag. To aid in generating such an environment, Gazebo provides a GUI alongside hundreds of object models (trees, rocks, buildings, etc) for creating complex indoor/outdoor terrain on which virtual robots can be placed. Once a robot model is placed and integrated with ROS, the robot can move around the virtual terrain and receive sensor data.

\textbf{NS-3.} Using the discrete event network simulator NS-3, it is possible to create an accurate prediction of how a network will perform given a specific network topology. The simulator describes each of the devices in the simulated network as nodes which each have their own network properties. It can then analyzing the distance, interference and individual properties between each of the nodes to predict network performance using metrics such as: packet loss probability and packet delay.

\textbf{Simulation environment design.} We generated simulated scenarios for both LOS and NLOS channels to demonstrate that our technique is capable of executing its intended function. The environments are rendered on Gazebo (\textit{Baylands} environment), which in our case, hosts the physics simulator. The designed environmental setting has dimensions of one hundred by one hundred meters, as seen in the figure \ref{simulation}. And each grid symbolizes a square with a side length of 20 meters. The environment was designed to accommodate both LOS and NLOS scenarios, and the majority of the environment is populated with trees, house model, wooden case, and elevated land to support the NLOS abstraction. A range of robotic agents, including UGVs and UAVs, have been chosen for use in this task. The UAV was chosen to be an Iris drone, while the UGV was chosen to be two Husky robots. We have used the MAVROS package for the connection between the Iris drone and the ROS environment. 
\begin{figure}[htbp]
\centering
\includegraphics[width=\linewidth]{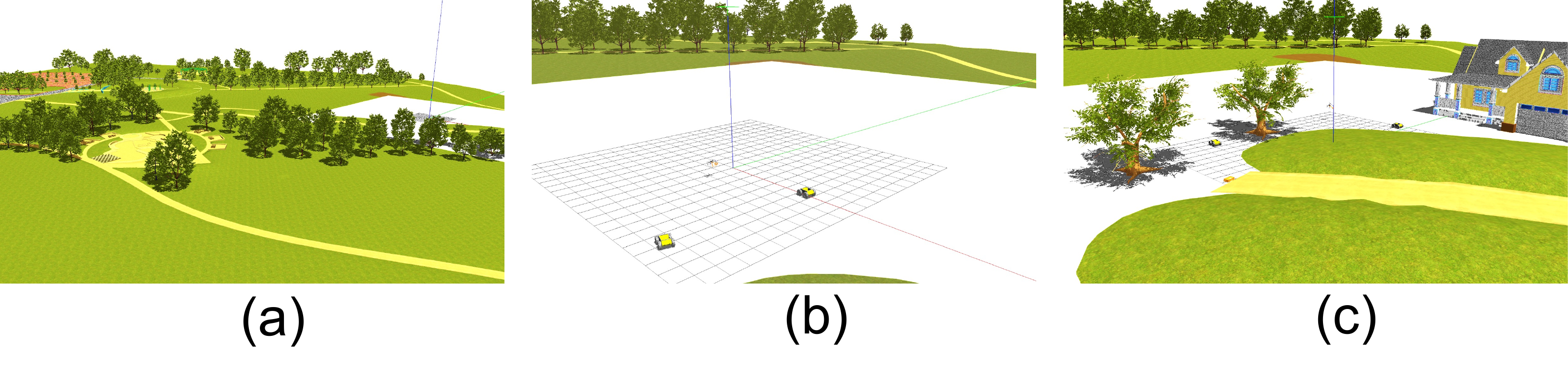}
\caption{The Gazebo simulation interface of our system. We have used (a) the Baylands terrain as our base environment. The next two images are (b) LOS, (c) NLOS communication scenario respectively. We have introduced trees, dense forest, wooden box, and house model for NLOS abstraction.}
\label{simulation}
\vspace{-1ex}
\end{figure}
Also, the data packets captured from the drone are published as MAVLINK messages. Considering the masterless scenario, each robotic agent is equipped with its own network. The IEEE 802.11 (Wi-Fi) interface has been used as our wireless communication medium. All the deployed agents are aware of the IPs of all the agents present in the simulation, and the agents themselves are configured to communicate data points across TCP/IP lines. We have captured ROSbag containing the accelerometric data of the robotic agents for a fixed 10 seconds every time we create a running event. In order to analyze the performance of the communication, we have opted to utilize one of the most advanced network simulators available, NS-3, which is also compatible with the wireless stack that we have chosen. In order to integrate Gazebo with NS-3, we deployed our technique as a synchronizing middleware. When employing our technique, the velocity difference between the agents were calculated from the data of the $/gazebo\_states/twist$ topic to adjust the window size. The velocity of the agents is modified in line with the circumstances in order to carry out this scenario. The algorithm changes the initial window size, which is set to 1 millisecond (mS).

The agents in the line of sight scenario are put in an environment devoid of any barriers, as shown in the figure \ref{simulation}. The ability to test communication performance while maintaining LOS abstraction is enabled by simulating the UGVs and the UAV traveling at different speeds. The performance matrices, according to the information in this section, include the average delay and the proportion of missed packets. The results are shown together with the resulting shift in the distance between the agents. For NLOS implementation, we have introduced dense forest and also elevated surfaces between agents when transmitting data.

\subsection{Simulation Results}
We investigate two unique communication scenarios: communication between UGVs and communication between UAVs and UGVs. The agents in the case of NLOS abstraction are programmed to operate in an object-populated environment. The signal's strength is likely to be reduced under these conditions. In this scenario, the reporting paradigm and communication conditions are identical to those used in the preceding case.
\begin{table}[h!]
\centering
\caption{Simulation results for our masterless synchronizing middleware on both LOS, and NLOS communication scheme using average delay ($PD_a$) (s) and packet loss probability ($L_p$) (\%) matrices considering a UGV to UGV communication scenario.}
\label{ugv}
\resizebox{\columnwidth}{!}{%
\begin{tabular}{|l|ll|ll|}
\hline
\multirow{2}{*}{\textbf{Distance (m)}} & \multicolumn{2}{l|}{\textbf{LOS}} & \multicolumn{2}{l|}{\textbf{NLOS}} \\ \cline{2-5} 
 & \multicolumn{1}{l|}{$PD_a$ (s)}  & $L_p$ (\%) & \multicolumn{1}{l|}{$PD_a$ (s)} & $L_p$ (\%) \\ \hline
20 & \multicolumn{1}{l|}{0.01} & 0 & \multicolumn{1}{l|}{0.2} & 0 \\ \hline
40 & \multicolumn{1}{l|}{0.1} & 0 & \multicolumn{1}{l|}{0.5} & 10 \\ \hline
60 & \multicolumn{1}{l|}{0.4} & 0 & \multicolumn{1}{l|}{0.8} & 10 \\ \hline
80 & \multicolumn{1}{l|}{0.4} & 10 & \multicolumn{1}{l|}{1} & 20 \\ \hline
100 & \multicolumn{1}{l|}{0.6} & 10 & \multicolumn{1}{l|}{1} & 30 \\ \hline
\end{tabular}%
}
\end{table}
When communicating between two UGVs, it has been shown that when the difference in relative velocity between the two vehicles is lower, the agents retain the data quantity during transmission for around 40 meters, as reported at table \ref{ugv}. In this case, the average delay (the sum of all delays experienced during the transmission of 10 seconds of accelerometric data) is also less than 0.1 second. Up to this point, the proposed alternative, which uses an adjustable window under masterless environment, perform well for line-of-sight communication. The master-based technique without window adjustment \ref{ros1}, on the other hand, begins to lose essential information significantly after this point, resulting in a delay. When the agents in this simulation are the furthest apart from one another, a relatively small adjustment to the sliding window value (0.1 milliseconds in this case) is observed to contribute to a nearly 20\% improvement in LOS retrieval and at least a 10\% improvement in NLOS retrieval (100m). 

Non-line-of-sight (NLOS) communication performance was significantly reduced during testing with UGVs to UAVs, as seen in the table \ref{uav}. To be more specific, when employing the masterless strategy with a distance of 100 meters or more separates each pair of agents, nearly all of the data is lost. This circumstance has an effect on the proposed adjustable window technique, which results in a 0.3 millisecond increase and leaves space for future advancement. Aside from this one issue, the redesigned window improves both line-of-sight (LOS) and non-line-of-sight (NLOS) communication significantly. In summary, we achieved an improvement of 40\% and 20\% in terms of average delay and packet loss probability respectively for NLOS scenario in comparison with master-based approach. For LOS, the margins are 20\% and 10\% respectively.
\begin{table}[h!]
\centering
\caption{Simulation results for our masterless synchronizing middleware on both LOS, and NLOS communication scheme using average delay ($PD_a$) (s) and packet loss probability ($L_p$) (\%) matrices considering a UGV to UAV communication scenario.}
\label{uav}
\resizebox{\columnwidth}{!}{%
\begin{tabular}{|l|ll|ll|}
\hline
\multirow{2}{*}{\textbf{Distance (m)}} & \multicolumn{2}{l|}{\textbf{LOS}} & \multicolumn{2}{l|}{\textbf{NLOS}} \\ \cline{2-5} 
 & \multicolumn{1}{l|}{$PD_a$ (s)} & $L_p$ (\%) & \multicolumn{1}{l|}{$PD_a$ (s)} & $L_p$ (\%) \\ \hline
20 & \multicolumn{1}{l|}{0.1} & 0 & \multicolumn{1}{l|}{0.5} & 0 \\ \hline
40 & \multicolumn{1}{l|}{0.2} & 10 & \multicolumn{1}{l|}{0.7} & 20 \\ \hline
60 & \multicolumn{1}{l|}{0.2} & 10 & \multicolumn{1}{l|}{0.8} & 20 \\ \hline
80 & \multicolumn{1}{l|}{0.5} & 20 & \multicolumn{1}{l|}{1.3} & 60 \\ \hline
100 & \multicolumn{1}{l|}{0.6} & 30 & \multicolumn{1}{l|}{1.6} & 90 \\ \hline
\end{tabular}%
}
\end{table}
\section{System Implementation} \label{system}
To compare the results from the simulated environment, we conducted experiments to emulate master (ROS1) and master-less (ROS2) architectures in the real world. However, the robotic agents are controlled manually and distances are far closer to avoid damaging equipment.

\subsection{Hardware}
The hardware used for this experiment are: two Turtlebot3 Burgers and a Duckiedrone.

\textbf{Duckiedrone.} is a UAV developed by Duckietown, a company known for its open source and differential robots. The drone is equipped with a Raspberry Pi 3b that we will use to produce the wifi signal, which can reach a distance of about 30 meters at 59/97.67 Mbps. Compared to higher end drones, the Duckiedrone has a significant amount of drift. ROS Kinetic and several other packages are installed on the drone to enable the use of positional control. This is where the drone automatically shifts its position to maintain a static image from the \textit{Arducam}, which is faced directly toward the ground.

\textbf{Turtlebot3 Burger} is a UGV developed by Robotis, which focuses on its compatibility with ROS and its available versions. This is exemplified by many ROS developers (such as the developers of \cite{Rosnetsim2021}) using the robot to demonstrate newly released packages. The robot is equipped with a LIDAR for object detection and two wheels controlled by an OpenCR board for movement. When the Turtlebot is activated it creates several rostopics, one of which is the $/scan$ topic that publishes scan information from the LIDAR. Since the LIDAR generates a large amount of data in a short amount of time, the $/scan$ topic will be used for ROSbag formation.

\subsection{Experimentation Flow}
The experimentation is categorized into two main sections: master-based and master-less approaches. Each approach is then tested using TCP and UDP in line-of-sight (LOS) and non-line-of-sight (NLOS) situations. The overall process can be summarized as follows:
\begin{enumerate}
    \item On the sender (Turtlebot) generate a ROSbag.
    \item Open filters to detect the amount of packets being sent and received. Then send the ROSbag.
    \item Use ping to determine the average delay.
    \item Repeat for an increased distance.
\end{enumerate}
This experimentation is repeated 20 times to generate an average packet loss and delay.

\begin{figure}[htbp]
\centering
\includegraphics[width=\linewidth]{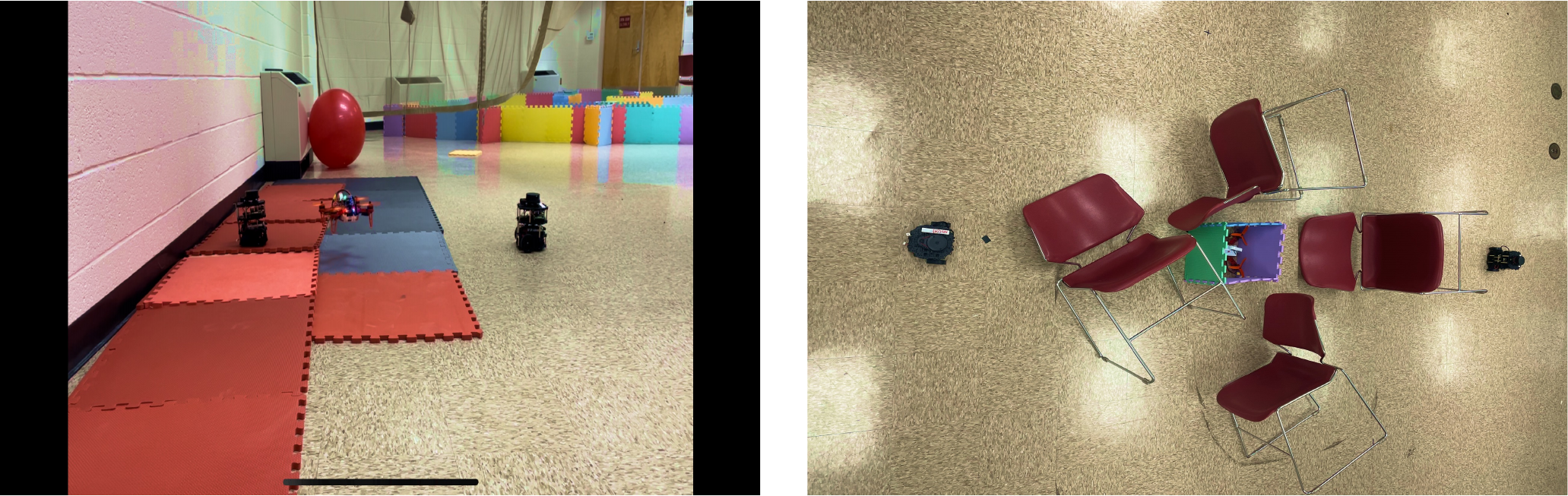}
\caption{Sample implementation of LOS (image on the left), and NLOS (image on the right) communication scenario with a Duckiedrone and two turtlebots.}
\label{system}
\vspace{-2ex}
\end{figure}
To Acheive LOS, Turtlebot(s) are placed in direct line-of-sight of the Duckiedrone. During NLOS we placed several chairs and a foam box obstructing line-of-sight. A sample illustration of the implementation scenario is provided in figure \ref{system}.
However, this scenario did not create any visible changes in the results. Therefore NLOS results are generated with the Turtlebot sender placed on the other side of a wall.

\subsection{ROSbag Formation}
The file to be sent across agents was generated by using ROSbag through subscribing to the /scan topic of a Turtlebot. The command is provided below:

\begin{lstlisting}[language=Python]
rosbag record -d 15 -O scan /scan
\end{lstlisting}

This command subscribes to the /scan topic of the Turtlebot for 15 seconds and saves the results to a file named $scan.bag$ in the current directory. In this experiment, the generated file reached a size of 293797 bytes.

\subsection{Communication Scheme}
For evaluating network performance, we used \textit{Netcat} traditional and \textit{TCPdump}. The following command scheme uses TCP to transfer the file. In order to use UDP $-u$ should be added after $nc.traditional$. A demonstration of the sample commands is given below:

\begin{lstlisting}[language=Python]
nc.traditional -l -p 1234 > recv.bag
tcpdump dst {recv_ip} and port 1234
\end{lstlisting}
The receiver initializes, \textit{Netcat} to begin listening on port 1234, any received data will be saved to a file named $recv.bag$. Then a filter is set for \textit{TCPdump} to capture any packets going to the receivers IP at port 1234. We then initiate the sender to send the file.

\begin{lstlisting}[language=Python]
tcpdump src {sender_ip} and port 1234
nc.traditional {recv_ip} 1234 < scan.bag
\end{lstlisting}

Similar to the receiver, a filter is set for \textit{TCPdump} to capture any packets coming from the $sender\_ip$ to a port 1234. \textit{Netcat} then sends $scan.bag$ to the receiver.

While conducting the experiment with the 293797 byte file. \textit{TCPdump} detected an average of 207, 502 byte packets sent using TCP. While for UDP an average of 36, 8192 byte packets were detected. The window size of each protocol will be used with ping, to approximate delay and time to send the file.

\subsection{Master (ROS1) based system}
To emulate a master based system the two Turtlebots connect to the Duckiedrones' network. In this way, all communication will pass through the Duckiedrone making it the master. We then begin the communication and testing procedure between the two Turtlebots.

\begin{figure}[htbp]
\centering
\includegraphics[width=\linewidth]{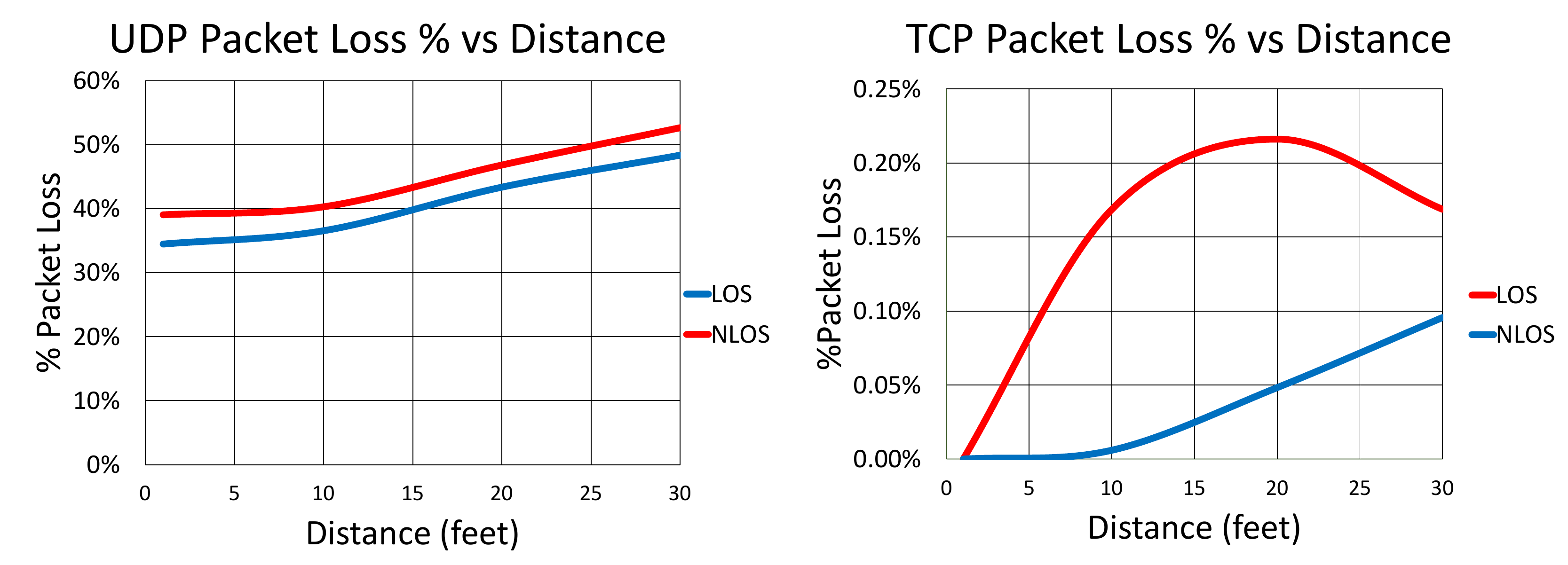}
\caption{A comparison of TCP and UDP is based on packet loss probability trends that vary with agent distance in a multi-agent environment. This system uses a master-based (ROS1) architecture.}
\label{experiment}
\end{figure}

UDP shows an initial packet loss of 35\%, which steadily increases as the robots are moved further away. When the sending Turtlebot is placed behind a wall, this shifts the packet loss by 5\%. At first glance of the TCP plot, it seems that NLOS is more efficient than LOS. However, the scale of the plots shows overall packet loss is less than 0.25\% meaning it is almost negligible. The detailed results can be found in figure \ref{experiment}.

Due to the close distances, with a maximum of 30 feet between each Turtlebot and Duckiedrone the average delay remains constant. In this architecture the delay would reach approximately 16ms for UDP (file transferred in 576ms) and 5ms for TCP (file transferred in 1035ms).

\subsection{Masterless (ROS2) based system}
In a Master-less system all robotic agents are connected to each other and communication doesn't flow through a single point. This testing represents a single branch in a master-less system using two robots as a full system . To emulate that branch, a Turtlebot connects to the Duckiedrone's network, and the procedure begins between the Turtlebot and Duckiedrone.

When completing the experiment up to a maximum distance of 55 feet between the Turtlebot and Duckiedrone. We could not detect any packet loss or change in packet delay when manipulating the environment. However, in contrast to the delay of the master based approach, the delay of UDP averaged at about 13ms (file transferred in 468ms) and the delay of TCP averaged at 4.5ms (file transferred in 931.5ms).

\section{Discussion} \label{discussion}
In this section we present some baseline studies with ROS1 implementation of our proposed synchronizing middleware and experimentation using both TCP and UDP transmission protocols. Also, we present the analysis on real-time network simulator, EMANE along with some possible future research directions.
\subsection{UDP Implementation}
We have experimented with ROS1 architecture and UDP connection to achieve the baseline performance. It is used as a comparative analysis with our proposed middleware. The steps of the UDP implementation for a ROS1 based synchronizing middleware is described below.

At every initialized socket we modified the sockets from using $SOCK\_STREAM$ to instead use $SOCK\_DGRAM$, which converts the sockets to UDP.
\begin{lstlisting}[language=Python]
sock = s.socket(s.AF_INET, s.SOCK_DGRAM)
\end{lstlisting}

As UDP sockets have a theoretical connection attribute, it can save the address of the sender. In order to simulate a connection we modify $recv\_all$ to return the address of the sender. Then modify $recv\_one\_message$ to connect to the address. The source code for this modification is listed below:
\begin{lstlisting}[language=Python]
#recvall
newbuf, address = sock.recvfrom(count)
return buf, address
#recv_one_message
lengthbuf, address = cls.recvall(sock, 4)
data, address = cls.recvall(sock, length)
sock.connect(address)
return data
\end{lstlisting}

Finally, whenever sending a UDP message it's important to separate the length of the buffer and message into two transmissions. UDP will push all the information from one message into the specified buffer size, even if the message requires more space. This is solved by converting $send\_one\_message$ in the following way.
\begin{lstlisting}[language=Python]
length = len(data)
sock.sendall(struct.pack('!I', length))
sock.sendall(data)
\end{lstlisting}
After completing the changes, we can run the example simulation and ping between two virtual devices.
\subsection{Varying transmission protocol within a master-based system}
We implemented our middleware on a master-based environment for both TCP and UDP to present a comparative analysis of our proposed masterless system with the ROS1 architecture. For both cases, the same Gazebo environment was chosen. After running the simulation, the results are shown in table \ref{ros1}. From LOS situations, we can observe that UDP communication has an overall higher packet loss probability but lower packet delay. However, at low distances, there is little to no difference between using UDP and TCP. So one thing to consider for the next simulation is to instead use a larger distance for better comparison of results. 
\begin{table}[h!]
\centering
\caption{Result comparison between TCP and UDP transmission protocol for master-based (ROS1) system considering both LOS and NLOS under a UGV to UGV communication scenario.}
\label{ros1}
\resizebox{\columnwidth}{!}{%
\begin{tabular}{|c|cccc|cccc|}
\hline
\multirow{4}{*}{\textbf{\begin{tabular}[c]{@{}c@{}}Distance \\ (m)\end{tabular}}} & \multicolumn{4}{c|}{\textbf{LOS}} & \multicolumn{4}{c|}{\textbf{NLOS}} \\ \cline{2-9} 
 & \multicolumn{2}{c|}{\textbf{\begin{tabular}[c]{@{}c@{}}Average   \\ Delay (s)\end{tabular}}} & \multicolumn{2}{c|}{\textbf{\begin{tabular}[c]{@{}c@{}}Packet Loss \\ Probability (\%)\end{tabular}}} & \multicolumn{2}{c|}{\textbf{\begin{tabular}[c]{@{}c@{}}Average \\ Delay (s)\end{tabular}}} & \multicolumn{2}{c|}{\textbf{\begin{tabular}[c]{@{}c@{}}Packet Loss \\ Probability (\%)\end{tabular}}} \\ \cline{2-9} 
 & \multicolumn{1}{c|}{\multirow{2}{*}{\textbf{UDP}}} & \multicolumn{1}{c|}{\multirow{2}{*}{\textbf{TCP}}} & \multicolumn{1}{c|}{\multirow{2}{*}{\textbf{UDP}}} & \multirow{2}{*}{\textbf{TCP}} & \multicolumn{1}{c|}{\multirow{2}{*}{\textbf{UDP}}} & \multicolumn{1}{c|}{\multirow{2}{*}{\textbf{TCP}}} & \multicolumn{1}{c|}{\multirow{2}{*}{\textbf{UDP}}} & \multirow{2}{*}{\textbf{TCP}} \\
 & \multicolumn{1}{c|}{} & \multicolumn{1}{c|}{} & \multicolumn{1}{c|}{} &  & \multicolumn{1}{c|}{} & \multicolumn{1}{c|}{} & \multicolumn{1}{c|}{} &  \\ \hline
\multirow{2}{*}{20} & \multicolumn{1}{c|}{\multirow{2}{*}{0.01}} & \multicolumn{1}{c|}{\multirow{2}{*}{0.01}} & \multicolumn{1}{c|}{\multirow{2}{*}{0}} & \multirow{2}{*}{0} & \multicolumn{1}{c|}{\multirow{2}{*}{0.18}} & \multicolumn{1}{c|}{\multirow{2}{*}{0.2}} & \multicolumn{1}{c|}{\multirow{2}{*}{10}} & \multirow{2}{*}{10} \\
 & \multicolumn{1}{c|}{} & \multicolumn{1}{c|}{} & \multicolumn{1}{c|}{} &  & \multicolumn{1}{c|}{} & \multicolumn{1}{c|}{} & \multicolumn{1}{c|}{} &  \\ \hline
\multirow{2}{*}{40} & \multicolumn{1}{c|}{\multirow{2}{*}{0.08}} & \multicolumn{1}{c|}{\multirow{2}{*}{0.1}} & \multicolumn{1}{c|}{\multirow{2}{*}{0}} & \multirow{2}{*}{0} & \multicolumn{1}{c|}{\multirow{2}{*}{0.4}} & \multicolumn{1}{c|}{\multirow{2}{*}{0.5}} & \multicolumn{1}{c|}{\multirow{2}{*}{30}} & \multirow{2}{*}{10} \\
 & \multicolumn{1}{c|}{} & \multicolumn{1}{c|}{} & \multicolumn{1}{c|}{} &  & \multicolumn{1}{c|}{} & \multicolumn{1}{c|}{} & \multicolumn{1}{c|}{} &  \\ \hline
\multirow{2}{*}{60} & \multicolumn{1}{c|}{\multirow{2}{*}{0.42}} & \multicolumn{1}{c|}{\multirow{2}{*}{0.48}} & \multicolumn{1}{c|}{\multirow{2}{*}{10}} & \multirow{2}{*}{10} & \multicolumn{1}{c|}{\multirow{2}{*}{0.54}} & \multicolumn{1}{c|}{\multirow{2}{*}{0.8}} & \multicolumn{1}{c|}{\multirow{2}{*}{50}} & \multirow{2}{*}{30} \\
 & \multicolumn{1}{c|}{} & \multicolumn{1}{c|}{} & \multicolumn{1}{c|}{} &  & \multicolumn{1}{c|}{} & \multicolumn{1}{c|}{} & \multicolumn{1}{c|}{} &  \\ \hline
\multirow{2}{*}{80} & \multicolumn{1}{c|}{\multirow{2}{*}{0.46}} & \multicolumn{1}{c|}{\multirow{2}{*}{0.6}} & \multicolumn{1}{c|}{\multirow{2}{*}{30}} & \multirow{2}{*}{10} & \multicolumn{1}{c|}{\multirow{2}{*}{0.96}} & \multicolumn{1}{c|}{\multirow{2}{*}{1}} & \multicolumn{1}{c|}{\multirow{2}{*}{60}} & \multirow{2}{*}{40} \\
 & \multicolumn{1}{c|}{} & \multicolumn{1}{c|}{} & \multicolumn{1}{c|}{} &  & \multicolumn{1}{c|}{} & \multicolumn{1}{c|}{} & \multicolumn{1}{c|}{} &  \\ \hline
\multirow{2}{*}{100} & \multicolumn{1}{c|}{\multirow{2}{*}{0.5}} & \multicolumn{1}{c|}{\multirow{2}{*}{0.8}} & \multicolumn{1}{c|}{\multirow{2}{*}{40}} & \multirow{2}{*}{20} & \multicolumn{1}{c|}{\multirow{2}{*}{1}} & \multicolumn{1}{c|}{\multirow{2}{*}{1.4}} & \multicolumn{1}{c|}{\multirow{2}{*}{60}} & \multirow{2}{*}{50} \\
 & \multicolumn{1}{c|}{} & \multicolumn{1}{c|}{} & \multicolumn{1}{c|}{} &  & \multicolumn{1}{c|}{} & \multicolumn{1}{c|}{} & \multicolumn{1}{c|}{} &  \\ \hline
\end{tabular}%
}
\end{table}
On the other hand, in NLOS situations, the results were somewhat self-explanatory. Similarly to LOS situations, there was little difference in packet delay, but, for the most part, delay was less in comparison to TCP. Packet loss probability, however, showed a more accurate comparison, showing that the packet loss probability for UDP was much higher than TCP. Another simulation to consider is testing a more congested NLOS environment, introducing more obstacles and observing their change to the results. Introducing obstacles between the agents would cause the aspects of UDP to increase. As TCP delay increases, UDP delay increases as well, albeit by a smaller amount. On the other hand, as packet loss in TCP increases, UDP packet loss also increases greatly. This means that when an NLOS environment is simulated for UDP, both the benefits of lower latency and the drawback of less reliability are made worse. To state the comparative margins between our technique and master-based: we improved average delay and packet loss probability by 40\% and 20\% for NLOS against the master-based technique. LOS margins are 20\% and 10\%, respectively.

\subsection{Experimentation with a real-time network simulator}
We have chosen the EMANE emulator to focus on real-time modeling of network configurations. This experimentation is conducted to counter the need for synchronizing middleware and check whether a real-time network emulator is sufficient to replace our proposed middleware along with the advantages we are offering. The integration of EMANE with ROS is still not officially included. As a result, it is not possible to replicate the same experimental setting as we did for our middleware. Instead, we have designed a similar data transmission environment inside of EMANE, utilizing its emulator mode to measure the same performance metrics we have used for our synchronizing middleware. 

We have designed a 10 node mesh network working on the TCP/IP transfer protocol. The default conditions create a 0\% packet loss with an average delay of 5.377ms using 100 pings of 64 bytes. With the increase in the longitude, latitude, and altitude of node 10 by 100, the packet delay increases to 54.309ms. This is with pinging from node 1 to node 10 without any hops.
The takeaway from this experiment is that although EMANE 
ensures the receipt of all the packets between nodes (TCP/IP mode), the packet delay is almost 10 times greater than our method. Moreover, the scenario was relatively less congested and NLOS was not present in EMANE. Creating the right ROS packages for integrating EMANE with a physics simulator could be an interesting area of research for analyzing masterless systems in depth.

\section{Conclusion} \label{sec:conclude}
In this work, we have questioned the communication overhead issue of the master-based ROS1 system while developing a synchronizing middleware for co-simulation of physics and network simulator. We have a middleware which works on a masterless system and is capable of handling the heterogeneity in a multi-robot environment. Additionally, to increase reliability in terms of minimizing packet loss probability, we have integrated a dynamic adjustment strategy within the TCP/IP sliding window. This algorithm can change the size of the window based on how fast the agents are moving so that they can work better together and communicate. We have designed extensive empirical analysis for both simulation and real robot deployment to showcase the efficacy of our proposed system. Even in difficult non-line-of-sight (NLOS) environments with heterogeneous agents, experimental data shows that our solution outperforms the standard fixed window-based strategy in terms of ensuring fewer packet losses (on average 10\% improvement) and faster transmission (about 12\% improvement). Experimental results also show that our synchronizing middleware can outperform the real-time network simulator, i.e., EMANE, in terms of efficient communication, which validates the necessity of such middleware.


\bibliographystyle{ieeetr}
\bibliography{main}
\end{document}